# A CRF Based POS Tagger for Code-mixed Indian Social Media Text

**Kamal Sarkar**
Computer Science & Engineering Dept.
Jadavpur University
Kolkata-700032, India
`jukamal2001@yahoo.com`

## Abstract

In this work, we describe a conditional random fields (CRF) based system for Part-Of-Speech (POS) tagging of code-mixed Indian social media text as part of our participation in the tool contest on POS tagging for code-mixed Indian social media text, held in conjunction with the 2016 International Conference on Natural Language Processing, IIT(BHU), India. We participated only in constrained mode contest for all three language pairs, Bengali-English, Hindi-English and Telegu-English. Our system achieves the overall average F1 score of 79.99, which is the highest overall average F1 score among all 16 systems participated in constrained mode contest.

## 1 Introduction

Part-of-Speech (POS) tagging POS tagging is an important preprocessing task in many NLP (Natural Language Processing) applications. Automatic analysis of non-standard texts like social media texts which differ from standard texts in the word usage and their grammatical structure creates the need for the adaption of POS tagging methods to such text types.

The ICON 2016 shared task on POS Tagging For Code-mixed Indian Social Media Text is focused on developing the POS tagger systems for code mixed Indian social media text for Bengali-English, Hindi-English and Telegu-English language pairs.

The earlier works on POS Tagging for standard texts used mainly the rule based models, supervised machine learning based models such as HMM (Hidden Markov Model) (Sarkar and Gayen, 2012; Sarkar and Gayen. 2013), CRF(Conditional Random Field) (Ekbal et.al., 2007), Support Vector Machines (SVM) (Ekbal et.al., 2008). A POS Tagging System of English-Hindi Code-Mixed Social Media Content has been presented in (Jamatia et.al., 2015). NLP tool contests, held in conjunction with ICON conferences, are the yearly events. The tool contest on POS tagging for code-mixed Indian social media texts was also an event in ICON 2015 (Sarkar, 2016).

Our proposed POS tagging system for social media texts, described in this paper, is developed based on Conditional Random Fields (CRF) trained using a rich feature set that includes contextual features, orthographic features, punctuation features and word length features.

## 2 CRF Based Model for POS Tagging

### 2.1 Conditional Random Fields

Conditional Random fields (CRFs) was originally introduced by Lafferty et.al. (2001), is a statistical sequence modeling framework for labeling and segmenting sequential data. CRFs calculate conditional probability distributions $P(y|x)$ of label sequence $y=(y_1, ...., y_n) \in \gamma^n$ given an observation sequence $x=(x_1, ..., x_n) \in \aleph^n$ as follows.

$$P(y|x) = \frac{\exp\sum_t(\sum_i \lambda_i f_i(x,y_t) + \sum_j \mu_j g_j(x,y_t,y_{t-1}))}{Z(x)}$$

$$Z(x) = \sum_y \exp\sum_t(\sum_i \lambda_i f_i(x,y_t) + \sum_j \mu_j g_j(x,y_t,y_{t-1}))$$

Where $Z(x)$ is normalization factor to ensure the proper probability, which can be obtained efficiently by dynamic programming; $g_j(x, y_t, y_{t-1})$ is a transition feature function of entire observation sequence $x$; $f_i(x, y_t)$ is the state feature function of observation sequence at state $y_i$. The model parameters $\lambda_i$ and $\mu_j$ are estimated from training data using advanced convex optimization techniques.

## 2.2 Feature Representation

The different features used for implementing our POS tagging system have been described in this section.

**Word context features:** We use the context of the current token as a feature. The context of the current token is defined by a window in which the current token is at the center of the window of size 5. We use four composite features by combining the words in the window: <prev. word, prev. of the prev. word>, <prev. word, current word>, <current word, next word> and <next word of current token, next of next word of the current token>.

**Language features**: The data released for ICON 2016 contest was also language-tagged. We observed that the language code column also contain some special tags such as "ne" for a named entity and the tag "univ" appears for the tokens like punctuations, tokens containing "@" or "#" etc. We used this information as a feature. We create a composite feature by combining language code and current token.

**Orthographic, punctuation and length features:** In this section we present a number of orthographic, punctuation and length features that we use.

*ContainsDigit:* Whether the current token contains any digit (binary feature).

*ContainsMoreDots:* Whether the current token contains more than one dot (.) (binary feature).

*ContainsSlash:* Whether the current token contains slash (\) (binary feature).

*ContainsMoreSlash:* If the current token contains more than one slash(\) (binary feature).

*ContainsAtTheRateBeg*: Whether the first character of the current token is '@' (binary feature).

*ContainsAtTheRate:* Whether the current token contains '@' anywhere in it (binary feature).

*VowelCount:* The value of this feature is computed by counting how many vowels are there in the current token (numeric feature).

*LongRepeatedCharSeqAtEnd:* Whether current token contains repeated character sequence at the end of the token (binary feature).

*ContainsLongVowelSeqInside:* Whether the current token contains a sequence of at least three vowels (binary feature).

*RemoveLongVowelSeqInsideBySingleVowel*: It is a string feature. Sometimes vowels are repeated to put emphasis on a word. This leads to a new word form for a word. For example, to give emphasis on the Bengali word "khub", it may be used as "Khuuuuuuub". We replace the sequence of repeated occurrences of a vowel in the current token by the vowel itself to create a new word form which is used as the feature value. If no such sequence is present in the current token, we used the current token itself as the feature value.

*ContainsDigitAndAlphabetBoth*: Whether the current token contains digits and alphabets both (binary feature).

*ContainsPureDigitSeq*: Whether the token contains all digits (binary feature).

*ContainsSeqOfSameChar:* Whether the current token is a repeated sequence of the same character (binary feature).

*ContainsAllCaps:* Whether all characters in the current token are the capital letters (binary feature).

*ThereExistsAsuffixDigitFollowsAlph:* Whether the current token contains a suffix starting with a digit and ending with alphabets (binary feature).

*ThereExistsAsuffixDigit6FollowsAlphabets:* Whether the current token has a suffix starting '6' and ending with alphabet sequence (binary).

*ContainsPuncSeq:* Whether the current token is punctuation sequence (binary feature).

*ContainsCharsOtherThanAlphDigitPunc:* Whether current token contains any character other than alphabets, digits and punctuations (binary feature).

*ContainsHash:* Whether the current token contains '#' (binary feature).

*ContainsHttp:* Whether the current token contains 'http' (binary feature).

*ContainsHyphen:* Whether the current token contains '-' (binary feature).

*ContainsColon:* Whether the current token contains ':' (binary feature).

*ContainsHyphenatedNumber:* Whether the current token is a hyphenated number (binary).

*NomalFormOfTheWord:* This is string feature. In the social media text, it is very common practice to write words in short, for example, "korte" may be written as "krte", which is usually formed by deleting inner vowels in the word, i.e. "korte" => "krte". To incorporate this issue, we maintain a word list for each language pair where each line in the list contains a short form of a word (vowel deleted form) and its original dictionary word. Given a social media word $w$, it is searched in the list and if found, the corresponding original word is retrieved and the retrieved word is used as the feature value, otherwise the feature value is set to $w$ itself.

*ContainsFirstPartAlphabetSecondPartContainsOtherThanAlphDigitPunc:* The value of this feature is set to 1 if the word has two parts and the first part is made by English alphabets and the second part contains some characters which are not English alphabets, digits or punctuations.

*Length:* We consider word length (wlen) as a feature and discretize it into 4 intervals to obtain five different features. The feature value is set to $L_{wlen}$ when wlen<=3, otherwise it is set to $L_4$.

*Prefix feature:* This is a string feature. If the length of the word, wlen >=2, P1 = prefix after deleting the last character else P1 = word. If *wlen* >= 3, P2 = prefix after deleting the last two characters else P2 = word. If *wlen* >= 4, P3 = prefix after deleting the last three characters else P3 = word. If *wlen* >= 5, P4 = prefix after deleting the last four characters else P4 = word.

*Suffix feature:* This is a string feature. If the length of the word, *wlen* >= 2, S1 = suffix containing the last character else S1 = word. If *wlen* >= 3, S2 = suffix containing the last two characters else S2 = word. If *wlen* >= 4, S3 = suffix containing the last three characters else S3 = word. If *wlen* >= 5, S4 = suffix containing the last four characters else S4 = word.

*Dynamic Feature:* The POS tag(s) of the previous word(s) is considered as the dynamic feature.

## 2.3 Training CRF

For implementing our proposed CRF based POS tagger, we have used the latest version: CRF++ 0.58 of the Open NLP CRF++ package[1]. We have designed the template file according to the features discussed in earlier sections. The training file and test file created by representing the training data and test data according to our defined feature set are submitted to CRF++.

## 3 EVALUATION AND RESULTS

The datasets released by the organizers of *ICON 2016 shared task on POS Tagging For Code-mixed Indian Social Media Text* (Jamatia and Das, 2016), contains three folders: one file for Bengali-English (BEN_EN) Language pair, one file for Hindi-English (HI_EN) language pair and one file for Telegu-English (TE_EN) language pair.

Under each language pair folder, there are two folders: one for coarse grained tag-set and another for fine grained tag-set. For each tag-set, there are three flies: one for facebook data, one for Tweet data and one for whatsapp data.

[1] https://taku910.github.io/crfpp/

| Lang-Pair | Total of sentences | | | | | | | | | | | |
|---|---|---|---|---|---|---|---|---|---|---|---|---|
| | Training data | | | | | | Test data | | | | | |
| | Coarse Grained | | | Fine Grained | | | Coarse Grained | | | Fine Grained | | |
| | FB | TWT | WA | FB | TWT | WA | FB | TWT | WA | FB | TWT | WA |
| Ben-Eng | 148 | 173 | 305 | 148 | 173 | 305 | 616 | 413 | 748 | 616 | 413 | 748 |
| Hin-Eng | 772 | 1097 | 764 | 772 | 1097 | 764 | 111 | 110 | 219 | 111 | 110 | 219 |
| Te-Eng | 744 | 744 | 494 | 744 | 744 | 494 | 247 | 249 | 198 | 247 | 249 | 198 |

Table 1. The description of the data for various language pairs, FB: Facebook, WA: whatsapp, TWT: Tweets

Each line in a training file contains tokens in the languages of concerned pair, Language tag and Part-of-Speech tag.

| Team | Language wise Average F1 Score | | | Overall F1 Avg. |
|---|---|---|---|---|
| | BN_EN | HI_EN | TE_EN | |
| Amrita_CEN | 75.67 | 80.08 | 79.31 | 78.35 |
| BITS_PILANI _TEAM 1 | 69.63 | 72.04 | 78.40 | 73.36 |
| BITS PILANI _TEAM 2 | 74.32 | 69.72 | 75.67 | 73.24 |
| BITS PILANI _TEAM 3 | 67.98 | 65.92 | 73.93 | 69.28 |
| MISIM-UB | 64.76 | 68.05 | 73.71 | 68.84 |
| Prakash Pimpale | 67.32 | 69.25 | 73.59 | 70.05 |
| PreCogTexts | 74.66 | 80.51 | 82.83 | 79.34 |
| IITP MNIT(Sys1) | 75.09 | 78.86 | 84.06 | 79.34 |
| IITP MNIT(Sys2) | 74.79 | 79.46 | 67.13 | 73.80 |
| Divya | | 77.61 | 65.06 | 71.34 |
| IIIT-H | 76.16 | 78.06 | 79.53 | 77.92 |
| **KS_JU** | **78.13** | **79.13** | **82.71** | **79.99** |
| NLP-NITMZ | 78.86 | 75.61 | 75.24 | 76.57 |
| Surukam | 73.48 | 81.02 | 75.89 | 76.80 |
| Anuj | ---- | 79.42 | ---- | 79.42 |
| Learner | 76.42 | 78.20 | 80.45 | 78.36 |

Table 2. Overall performance of the participating systems in terms of average F1 score.

Each line in a test file contains a token in the languages of concerned pair and its language tag.

For each language pair, we develop two different CRF models- one for coarse grained tag-set and another for fine-grained tag-set.

For a given tag-set and a language pair, we combine all three available training files-one for facebook data, one for tweet data and one for whatsapp data to create a common training data set which is used to train the model. With this trained model, we test the system separately on the facebook data, tweet data and whatsapp data supplied as test sets for the corresponding language pair and the tag-set. The description of the data is shown in Table 1.

Our developed POS system has been evaluated using the traditional precision, recall and F1-measure.

For each language pair the contests were done in two different modes: Constrained mode and unconstrained mode. In constrained mode, the participant team is allowed to use the training corpus only. In unconstrained mode, the participant team is allowed to use any external resources to train their system. We have only participated in the constrained mode.

| Team Name | Language Wise Rank | | | Avg. Rank Score |
|---|---|---|---|---|
| | BN-EN | HI-EN | TE-EN | |
| Amrita_CEN | 5 | 3 | 6 | 4.67 |
| PreCogTexts | 8 | 2 | 2 | 4 |
| IITP_MNIT(sys1) | 6 | 7 | 1 | 4.67 |
| KS_JU | 2 | 6 | 3 | 3.67 |
| Learner | 3 | 8 | 4 | 5 |

Table 3: Comparisons of the participating systems obtaining avg. rank value of 5 or less (The systems obtaining avg. rank value of more than 5 are not shown)

The results of the participating systems were calculated in terms of precision, recall and F1-measure, but the official rank order of the systems was not published. So, we have analyzed the results in two angles: (1) we have in Table 2 the overall performance of a system in terms of average F1 score which is computed over F1 scores obtained by the system for all types of test data (facebook, Twitter and whatsapp) across all three language pairs and (2) we have shown in Table 3 the systems obtaining avg. rank value of 5 or less. The average rank value of a system is computed by averaging the language-pair wise ranks obtained by the system.

The team- IITP_MNIT submitted two systems in constrained mode contest. We mention them in the tables as IITP_MNIT (Sys1) and IITP_MNIT (Sys2). As we can see from Table 2, our system (team: KS_JU) obtains the average F1 score of 79.99, which is the highest among all 16 participating systems. Table 3 shows that our system also obtains the average rank value of 3.67, which is the best among all 16 participating systems.

## 4 Conclusion

This paper describes a CRF based POS tagging system for code mixed social media text in Indian Languages. Our system performs well across all three language pairs. We hope that the proper choice of features along with the suitable combination of machine learning algorithms would improve performance of our system.